\documentclass[10pt, a4paper]{article}

\usepackage{lrec-coling2024} 
\usepackage{listings}
\usepackage{multirow}
\usepackage{latexsym}
\usepackage{natbib}
\usepackage{multibib}
\usepackage{amsmath}
\usepackage{textcomp}

\usepackage{graphicx}
\usepackage{tabularx}
\usepackage{soul}
\usepackage{dirtytalk}  

\usepackage{booktabs}
\usepackage{xcolor}
\usepackage{hyperref}
 \definecolor{ForestGreen}{rgb}{0.14, 0.539, 0.24}
 \definecolor{darkblue}{rgb}{0, 0, 0.5}
  \hypersetup{colorlinks=true, citecolor=darkblue, linkcolor=darkblue, urlcolor=darkblue}

\usepackage{xstring}

\usepackage{color}

\newcommand{\bivert}[0]{\textcolor{ForestGreen}{\textsc{BiVert}}}

\DeclareMathOperator{\len}{len}

\title{\bivert: Bidirectional Vocabulary Evaluation using Relations for Machine Translation}

\name{Carinne Cherf \quad Yuval Pinter} 

\address{Department of Computer Science, Ben Gurion University \\ Beer Sheva, Israel \\
         \texttt{carinnecherf@gmail.com} \quad \texttt{uvp@cs.bgu.ac.il}\\}

\abstract{
Neural machine translation (NMT) has progressed rapidly in the past few years, promising improvements and quality
translations for different languages. Evaluation of this task is crucial to determine the quality of the translation.
Overall, insufficient emphasis is placed on the actual sense of the translation in traditional methods.
We propose a bidirectional semantic-based evaluation method designed to assess the sense distance of the translation from the source text.
This approach employs the comprehensive multilingual encyclopedic dictionary BabelNet.
Through the calculation of the semantic distance between the source and its back translation of the output, our method introduces a quantifiable approach that empowers sentence comparison on the same linguistic level.
Factual analysis shows a strong correlation between the average evaluation scores generated by our method and the human assessments across various machine translation systems for English-German language pair.
Finally, our method proposes a new multilingual approach to rank MT systems without the need for parallel corpora.
 \\ \newline  
}

\begin{document}

\maketitleabstract

\section{Introduction}
Automatic evaluation of machine translation (MT) is crucial to determine the quality and performance of translation systems.
It is an important step in the development and improvement of MT models, as it sheds light on the models' strengths and weaknesses.
As the demand expands for high-quality translations, spanning a variety of languages, also the need for efficient and reliable evaluation techniques grows rapidly.
The major goal of these evaluation methods is to approximate the semantic similarity between the target text and some generated text.
Standard techniques rely on comparing the machine translation’s output with the desired true reference. Common methods such as BLEU~\cite{papineni-etal-2002-bleu} and ROUGE~\cite{lin-2004-rouge} rate the translation based on n-gram intersections.
Many of these methods are effective at capturing aspects of text similarity, but fall short on the actual meaning difference.
Advanced techniques using word-embedding based approaches like BERTScore~\cite{zhang2020bertscore} have marked significant progress in assessing translation quality.
With this in mind, a significant shift has occurred in recent years towards the need for accurate reference-less evaluation metrics. 
\begin{figure*}[!ht]
\begin{center}
\includegraphics{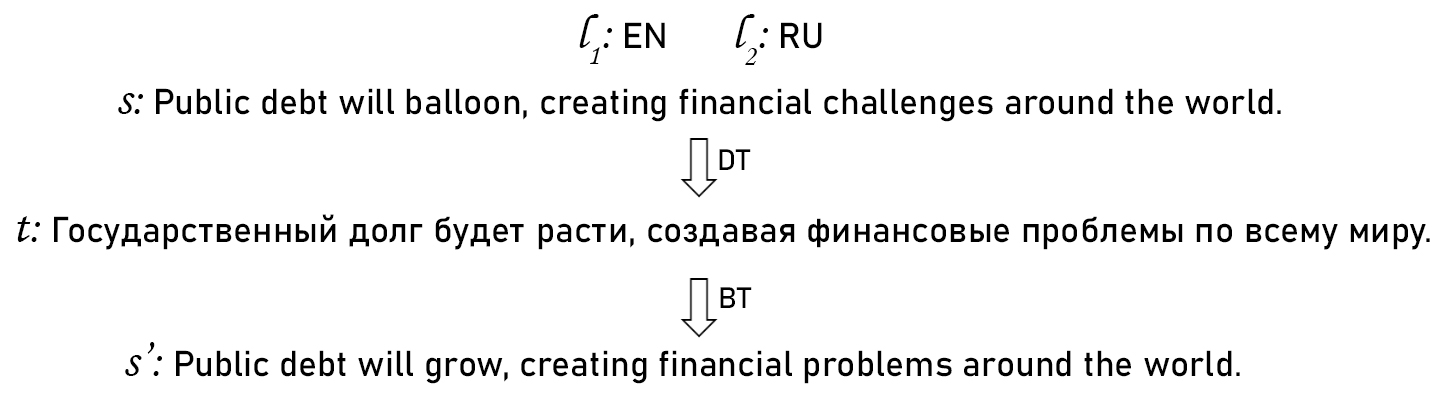} 
\caption{Example of a direct translation from English to Russian using the system we wish to evaluate, and its back-translation using a state-of-the-art translation system suitable for \bivert.}
\label{fig:directions}
\end{center}
\end{figure*}

Our goal is to introduce a different strategy for machine translation evaluation, one that does not require an aligned parallel test set.
\bivert{} is a simple bidirectional and self-supervised method constructed from a multilingual encyclopedia. 
In essence, \bivert{} evaluates a translation between the source sentence $s$ and the target sentence $t$ by scoring the semantic similarity between the $s$ and its back-translated sentence $s'$, as illustrated in \autoref{fig:directions}.
We refer to the former translation as the \textit{direct} action and the latter as the \textit{back} action. 
For the first step, we generate the back sentence $s'$ using a standard machine translation system, which we commonly label as a state-of-the-art MT system.
This way we form a single-language platform for comparing the meanings between the original text and the back translated text.
With the help of contextualized embeddings, extracted by the model to be evaluated, we pair the words between the sentences and compare them. At this point, we can estimate the semantic distance between $s$ and $s'$ making use of the word pairs, resulting in an indirect estimation of the direct translation quality.
We train \bivert{} features on the WMT Metrics Task 2021 dataset, and experiment on the WMT Metrics Task 2022 dataset, comparing our average results to existing methods~\cite{freitag-etal-2022-results}.
Our experiments show that \bivert{} obtains strong correlation with the human scores for the English--German language pair, with promising potential on Chinese to English and English to Russian.

\section{Related Work}

Numerous methods measure the resemblance between generated text and human text such as classic n-grams techniques and word embeddings strategies, some of which rely on a predefined reference. Previous research findings~\cite{novikova-etal-2017-need} cast doubt on the alignment between predicted outcomes and human judgments for known methods.
Recent advancements in the field of quality estimation have introduced techniques that offer a more accessible solution as they do not require collecting human references or obtaining parallel alignments.
Moreover, Previous research~\cite{Atranslationalbasisforsemantics} introduced a knowledge discovery technique known as Semantic Mirroring, which relies on identifying semantic relationships between words in a source language and their counterparts in a target language. They emphasize that by mirroring source words and target words back and forth they are able to provide insights into cross-lingual semantic relations.

\paragraph{Reference-based} measures assess the output of an MT system by comparing it to a limited set of reference text samples.
Traditional methods, such as BLEU and ROUGE which search for matching n-grams, primarily aim to capture prominent similarities between the generated text and the true reference.
To compare generated data against human text, Self-BLEU~\cite{ZhuYaoming2018Texygen} treats one sentence as a hypothesis and those remaining as references.
It calculates the BLEU score for each generated sentence in comparison to the collection, as the average BLEU score is then defined as the document's Self-BLEU mark.
Moreover, BERTScore~\cite{zhang2020bertscore}, an advanced evaluation technique, measures the similarity of two sentences as the sum of their cosine similarities between their pre-trained BERT contextual embeddings~\cite{devlin-etal-2019-bert}.
Although contextual embeddings are trained to capture long-range relationships effectively, they can still struggle with distinguishing between similar senses or meanings.
BERTScore is affected by the antonymy problem~\cite{saadany-orasan-2021-bleu}, where antonyms usually have similar contextual values and are closer in vector space.
As a result, a translation of one word to its exact opposite is not sufficiently captured as erroneous by the metric.
Another issue is that BERTScore struggles to distinguish between the mistranslation of a critical word that could significantly alter the intended meaning.
Occasionally, a word may have multiple interpretations depending on the context, whereas BERTScore may fail to capture the error that affects the actual sentence intention.
However, MoverScore~\cite{zhao-etal-2019-moverscore} takes into account the Euclidean distances between the vector representations and tries to find the minimum effort to transform between both texts. This captures more effectively the degree of resemblance between the texts. 
An alternative approach, MAUVE~\cite{pillutla2021mauve}, compares characteristics of the source and the target distributions using the Kullback-Leibler (KL) method.
It creates a divergence curve that represents two types of errors: false positives (unlikely text) and false negatives (missing plausible text).
By analyzing this curve and calculating the area under it, MAUVE provides a scalar value that quantifies the overall gap between both texts.
We note that although evaluation of individual sentence-level texts against references is beneficial, corpus-based metrics provide a more comprehensive and meaningful assessment of machine translation systems.

\paragraph{Quality estimation} (QE) for machine translation, also known as reference-less evaluation, presents an approach for assessing text, in particular relevant for authentic text, such as social media.
Moreover, it can also drastically decrease the cost of developing effective machine translation systems.
These methods value the quality of the translation without any information about aligned referenced text.
For instance, CometKiwi~\cite{rei-etal-2022-cometkiwi} implements this manner by combining qualities of two frameworks, Comet~\cite{ReiStewart2020COMET} for the training process and OpenKiwi~\cite{kepler-etal-2019-openkiwi} for prediction.
Their architecture feeds a trained network with both the source and target sentence resulting a score for the task, thus not requiring a reference text for the evaluation.
DeepQuest~\cite{ive-etal-2018-deepquest}, a sophisticated neural-based sentence-level architecture for document-level quality estimation, achieves impressive performance compared to previous methods.
The results of quality estimation can be either represented by standard metrics like F-measure or by determining the correlation between the evaluation score and the state-of-the-art gold standard.
In contrast, \bivert{} does not require a neural network training, as it is based on a multilingual connected sense-based network of words and only requires tuning of seven parameters.

\begin{figure*}[!ht]
\begin{center}
\includegraphics[scale=1.5]{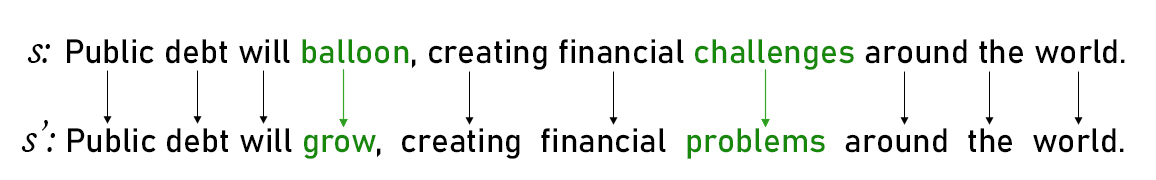} 
\caption{An example of final words alignment using the linear sum assignment problem algorithm.  }
\label{fig:align}
\end{center}
\end{figure*}

\paragraph{Semantic Graphs} provide a structured illustration of relationships between associated objects.
These graphs represent a network of words and senses, connected based on a relationship between both sides.
Word-sense disambiguation (WSD), a task of identifying the accurate sense of a word within a context, can be approached through graph-based algorithms.
Many words have multiple senses, and the challenge of determining the correct sense of a word often relies on the surrounding context.
In WSD, given a document represented as a sequence of words $W = \{w_1, w_2, ..., w_n\}$, the goal is to establish connections with the correct sense(s) for $w_i \in W$. Specifically, the objective is to find a mapping $f$ from the searched words to their senses, such that $f(w_i;W) \in S(w_i)$, where $S(w_i)$ is the set of senses for the word $w_i \in W$.
By forming semantic graphs assembled from words as nodes connected by edges representing semantic relationships, graph-based algorithms can resolve the obscure puzzle of connections between words.
WordNet~\cite{miller-1992-wordnet} is a prime example of semantic graphs, being a comprehensive lexical database that bridges semantic relationships among different concepts.
Various approaches such as MetaGraph2Vec~\cite{zhang2018metagraph2vec} and Edge2vec~\cite{WangChan2020Edge2Vec} benefit from sense networks for learning embeddings.

\begin{figure*}[!ht]
\begin{center}
\includegraphics[scale=0.39]{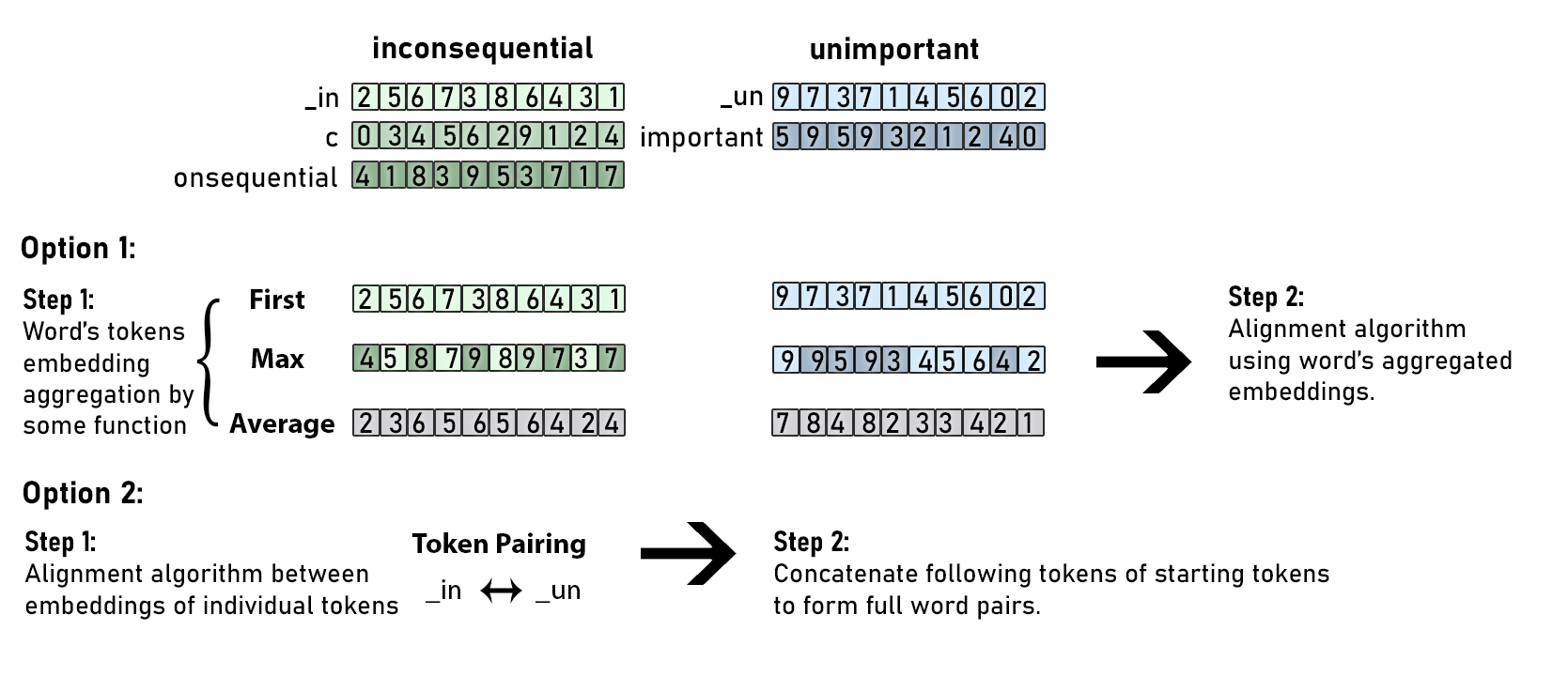} 
\caption{Example of words \textit{inconsequential} and \textit{unimportant} with illustrative embedding values, demonstrating different subword pooling strategies for word alignment. The word alignment algorithm calculates the cosine similarity between the embeddings representing the words chosen via option 1 or 2.}
\label{fig:tokens}
\end{center}
\end{figure*}

\section{\bivert: A Semantic Evaluation}
\label{sec:model}
\bivert{}, or \textcolor{ForestGreen}{\textbf{Bi}}directional \textcolor{ForestGreen}{\textbf{V}}ocabulary \textcolor{ForestGreen}{\textbf{E}}valuation using \textcolor{ForestGreen}{\textbf{R}}elations for machine \textcolor{ForestGreen}{\textbf{T}}ranslation, is an evaluation method for multilingual translation that concentrates on identifying the actual senses of the source sentence $s$ and the target sentence $t$.
This is achieved through comparing the source sentence $s$ and its back-translated sentence $s'$, both of whom share a common language $l_1$, allowing to calculate the semantic distance between them using only monolingual resources.
The first step is to generate the back-translated sentence $s'$ using a state-of-the-art translation system. An alternative use case could be to rely on the evaluated system itself for the back-translation. Any translation system of adequate quality can be employed for this task.
This is followed by matching word pairs between both sentences using a pairing algorithm on the words embeddings. The words embeddings might be split by sub-words and need to be aggregated.
We then identify the relation of each pair and assign a score accordingly (see section~\ref{ssec:relations}).
We sum the scores achieved by the word pairs for each category.
Finally, the assessment of the translation's quality is accomplished by aggregating the summed scores of all categories using trained weights for each relation type, which are tuned for each language pair, as detailed in section \ref{ssec:score}. 

\subsection{BabelNet}
One of \bivert{}'s objectives is to identify the correct sense connection between a pair of words. 
To this end, we make use of BabelNet~\cite{NAVIGLI2012217}, a consistently updated multilingual encyclopedic dictionary that connects named entities in a very large network of semantic relations.
BabelNet follows the WordNet model, consisting of \textit{synsets}, each representing a set of synonyms which encode the same concept.
Synsets are linked to each other using semantic relation edges of types such as \textit{hypernym}, \textit{hyponym}, and \textit{antonym}.
BabelNet is unique in providing extensive coverage of words and their meanings across multiple languages.
Moreover, BabelNet aggregates data from a variety of resources: Wikipedia, Wiktionary, Wikidata, VerbAtlas, WordNet, GeoNames and OmegaWiki. 

\subsection{Word Alignment}
Following the action of back-translating the target sentence $t$ into $s'$, we proceed to align the words between $s$ and $s'$, thereby generating pairs of matching words as demonstrated in \autoref{fig:align}.
To ensure accurate alignment of word pairs, we calculate the cosine similarity score between the embeddings corresponding to the aligned elements in both sentences. We match element pairs using the linear sum assignment problem (LSAP), implemented using a modified Jonker-Volgenant algorithm~\cite{CrouseDavid2016LSAP}.
LSAP is equivalent to minimum weight matching problem in bipartite graphs.
The objective is to pair each row with a distinct column in a manner that minimizes the sum of the corresponding entries.
In other words, we want to select $n$ tokens (rows) from $s$ and find their corresponding matches (columns) in $s'$ while maximizing the sum of cosine similarities.
Since the systems we evaluate on and with employ subword token embeddings, we require a way for pooling multiple tokens that correspond to a single word when such a segmentation occurs.
In one approach, the overall sentence-level alignment is performed over the token sequence, obviating the need for word-level aggregation.
Other methods encourage subword pooling as a preliminary step for word-level operations~\cite{acs-etal-2021-subword}
For instance, the maximum element-wise approach aggregates the tokens embeddings into single word representations by selecting the maximum value at each position of the embedding.
Another strategy settles on the first token for the word representation.
We chose to operate over the token level, selecting word alignments based on tokens they contain as \say{representatives} for the full word: as soon as a token inside a word is aligned, the word in its entirety is paired with the corresponding token's word from the other sequence. Follow the example in \autoref{fig:tokens}.

\subsection{Word Pair Relations}
\label{ssec:relations}
After pairing the words from the source and backtranslated text we define each pair's relationship.
We identify the following categories of possible word relations: \textit{Same}, \textit{Extra}, \textit{Missing}, \textit{Stopwords}, \textit{Inflection}, \textit{Derivation}, and \textit{Sense}.
Each match receives a value according to its type as described below.

\begin{enumerate}
  \item \textbf{Same}: This category refers to word pairs in which both words are identical. Since this pair does not cause any variation between the sentences, it is not taken into account within the final score decision. Their presence does not affect the evaluation hence the score assigned is zero.

  \item \textbf{Extra}: The extra category suggests a word has been added to the translation sentence and has no match in the source counterpart. This relation costs $1/\len(s)$, to account for its relative a-priori weight in the sentence.

  \item \textbf{Missing}: Missing word pair indicates a word from the source sentence $s$ lacks a parallel match in the back-translated sentence $s'$. A missing pair costs $1/\len(s)$ as well.
  
  \item \textbf{Stop words}: Non-identical paired words which are both contained in a list of language-specific stopwords are treated as one half of a replacement operation and cost $1/len(s)$, since they are often interchangeable (e.g., `at' $\leftrightarrow$ `on').
  
  \item \textbf{Inflection}:
  Inflection refers to a process of word formation to signal differences in grammatical attributes like tense, person, number, and gender. Two words are categorized as an inflection if their lemmas are identical. We weigh this relation by calculating the cosine similarity between both words' embeddings.
  
  \item \textbf{Derivation}: Derivation is the process of varying a word's part of speech while retaining its core semantic content. For instance, \say{happy} and \say{happiness} have a derivation relationship. We assess these pairs by computing their cosine similarity.
  
  \item \textbf{Sense}: Sense-related words are different words which have been chosen by the alignment algorithm due to their close embedding distance. These words may be synonyms, hypernyms, or antonyms. We aim to grade the actual distance of their intentional sense in the given context, using the multilingual encyclopedia BabelNet. For this issue we assemble a semantic graph described in section~\ref{ssec:senseRelation}.
\end{enumerate}

\begin{figure*}[!ht]
\begin{center}
\includegraphics[scale=0.35]{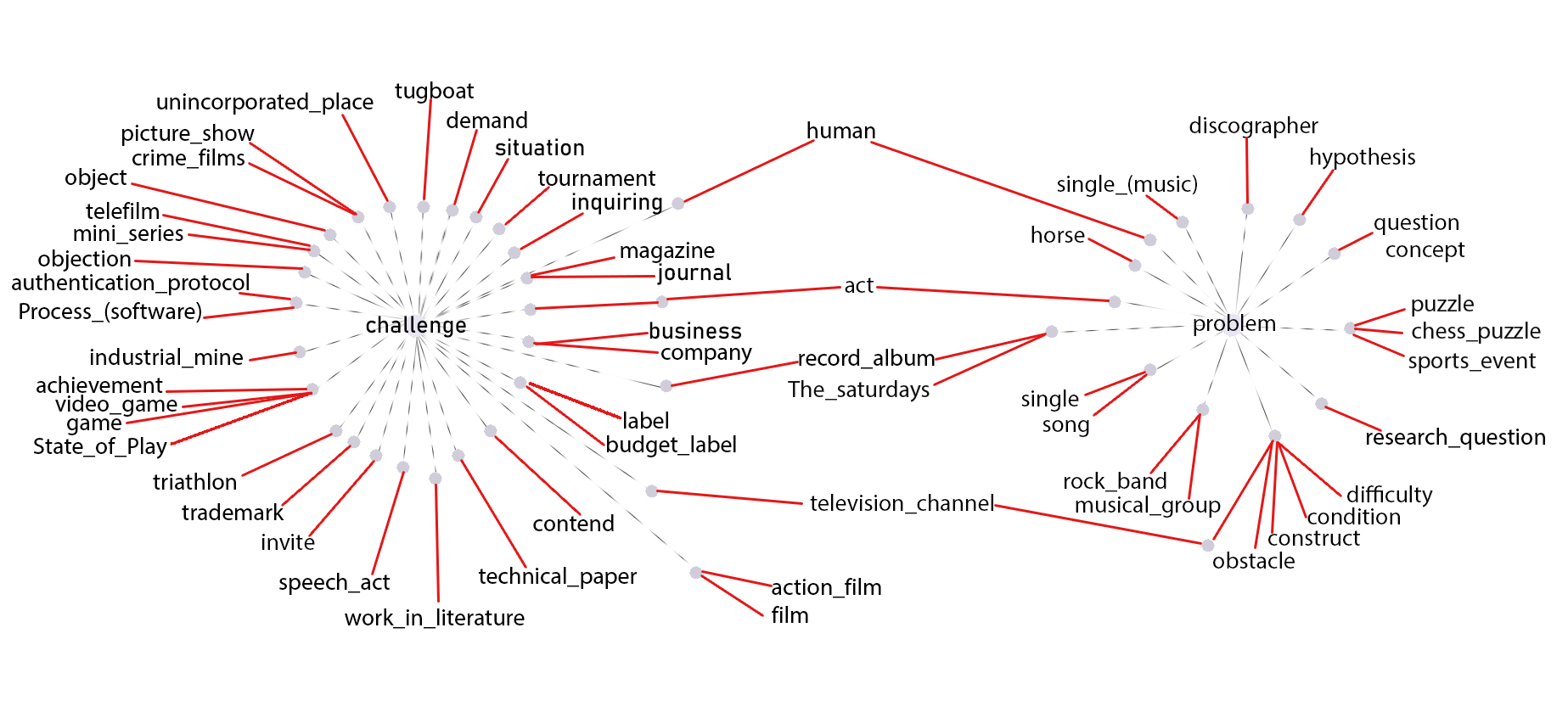} 
\caption{Fragment of a semantic graph between the two words \textit{challenge} and \textit{problem}. The hatched grey edges connect roots to their senses, and the red edges represent hypernym relations between the nodes contents.}
\label{fig:graph}
\end{center}
\end{figure*}

\subsubsection{Sense Relation Type}
\label{ssec:senseRelation}
The \bivert{} evaluation method is focused on finding the differences between words' true senses in order to correctly estimate the direct translation.
For each word pair found to exemplify the \textit{sense} relation, we form a semantic subgraph using BabelNet.
To construct the graph we pass both words, $x \in s$ and $y \in s'$, through a lemmatizer, if available in language $l_1$, and extract their senses.
The graph now has two roots, $x$ and $y$, and nodes connected to each root representing their senses.
We locate the shortest path from root $x$ and root $y$ using Dijkstra's algorithm~\cite{dijkstra1959note}.
As long as a path between the two roots has not been found, we continue expanding the graph by extracting each sense's hypernyms and iteratively searching for a connected path, as illustrated in \autoref{fig:graph}.
After marking the route, we score it as described in the remainder of the section.
If a path is not found according to a pre-specified max search depth threshold, we revert to scoring the relation as the cosine similarity between the roots.
We note that BabelNet's resources restrict us to scoring relations between nouns and between verbs.

The \textbf{sense score} for a matching pair is calculated using the semantic graph $G$, constructed from nodes $V$ representing the root words and their senses, and edges $E$ consistent of the relations between the nodes.
Each edge receives a score by the type of lexical connection it represents according to research done by Michael Sussna~\cite{WSDSussnaUsingNetwork1993}.
Each edge weight consists of type weights defined by the relation of the words~(\ref{eq:edge-weight}).
The type weight~(\ref{eq:edge-type-weight}) is defined by minimum and maximum values chosen for word relations of types hypernymy, hyponymy, holonymy, and meronymy.
In practice, all of these relations have weights ranging from 1 to 2.
In contrast, the weight used for all antonymy arches is constantly valued at 2.5.
The edge weight is then averaged by the two inverse weights and divided by the depth of the edge within the graph.
Together, the weight between node $a$ and $b$ is defined as:
\begin{equation}
\label{eq:edge-weight}
    w(a, b)=\frac{w\left(a \rightarrow_{r} b\right)+w\left(b \rightarrow_{r^{-1}} a\right)}{2 d},
\end{equation}
\begin{equation}
\label{eq:edge-type-weight}
    \quad w\left(x \rightarrow_r y\right)=\max_r-\frac{\max_r-\min_r}{n_r(X)},
\end{equation}
where $\rightarrow_r$ is a relation of type $r$ and $r^{-1}$ is its inverse;
$d$ is the depth of the deeper of the two nodes; $\max_r$ and $\min_r$ are the maximum and minimum weights possible for a relation of type $r$; and $n_r(X)$ is the number of relations of type $r$ leaving node $\mathrm{X}$.

The final graph score $S(a, b)$ from root $a$ to $b$ is given by the normalized sum of the edge weights along the path between them:
\begin{equation}
    S(a, b)=2\times(0.5-\frac{1}{\sum_{e \in P(a\leadsto b)} w(e)}).
\end{equation}

\begin{table*}[!ht]
\centering
\begin{tabular}{lcccccc}
    \toprule 
     & \textbf{Extra} & \textbf{Missing} & \textbf{Stopword} & \textbf{Inflection}& \textbf{Derivation} & \textbf{Sense} \\
    \midrule
    \multirow{1}{*}{English-German} & 0.121 & 0.134 & 0.188 & 0.101 & 0.092 & 0.360 \\
    \multirow{1}{*}{English-Russian} & 0.112 & 0.164 & 0.196 & 0.087 & 0.063 & 0.375 \\
    \multirow{1}{*}{Chinese-English} & 0.172 & 0.203 & 0.126 & 0.000 & 0.000 & 0.497 \\
    \bottomrule
\end{tabular}
\caption{\label{tbl:importances} Feature importance scores learned by a Gradient Boosting Regression model for \bivert{} language pairs. }
\end{table*}

\begin{table*}[!ht]
\centering
\begin{tabular}{lccccc}
    \toprule 
    Language pair & \textbf{eng-deu} & \textbf{eng-deu} & \textbf{eng-rus} & \textbf{zho-eng}& \textbf{zho-eng} \\
    Human Translation Included & yes & no & no & yes & no \\
    \midrule
    BERTScore & 0.338 & 0.428 & 0.811 & 0.843 & 0.924 \\
    Cross-QE & \textbf{0.643} & 0.661 & \textbf{0.806} & \textbf{0.817} & 0.870 \\
    COMETKiwi & 0.592 & \textbf{0.674} & 0.763 & 0.795 & 0.866 \\
    MS-COMET-QE-22 & 0.417 & 0.539 & 0.672 & 0.799 & \textbf{0.897} \\
    UniTE-src & 0.509 & 0.509 & 0.779 & 0.791 & 0.874 \\   
    MATESE-QE & 0.363 & 0.337 & 0.637 & 0.741 & 0.767 \\
    COMET-QE & 0.480 & 0.502 & 0.468 & 0.544 & 0.569 \\
    KG-BERTScore & 0.369 & 0.400 & 0.612 & 0.617 & 0.743 \\
    HWTSC-TLM & 0.311 & 0.428 & 0.597 & 0.368 & 0.460 \\
    HWTSC-Teacher-Sim & 0.290 & 0.385 & 0.675 & 0.294 & 0.356 \\
    \textbf{\bivert{}} & \textbf{0.694} & \textbf{0.703} & 0.657 & 0.376 & 0.239\\
    \bottomrule
\end{tabular}
\caption{\label{tbl:corr} System-level Pearson correlation between human scores and \bivert{} scores, compared to other evaluation metrics.  ``Human Translation Included'' refers to refB system which may be included or excluded from the correlation calculation. See system-level scores in \autoref{tab:all-scores}. Highest reference-free scores are \textbf{bolded}.}
\end{table*}

\subsection{Final Score}
\label{ssec:score}
Training and testing evaluation strategies occasionally requires human interference with the desired output.
Our procedure does not require any human resources as our method is completely automatic, comparing the source sentence with the generated backtranslated sentence.
The score of each relation pair is summed by relation categories.
The final score of \bivert{} is a trained combination of all relation types into a final score.
We use gradient descent to train our method in order to achieve optimal predictions for each language pair.

\begin{figure*}[!ht]
\begin{center}
\includegraphics[scale=0.6]{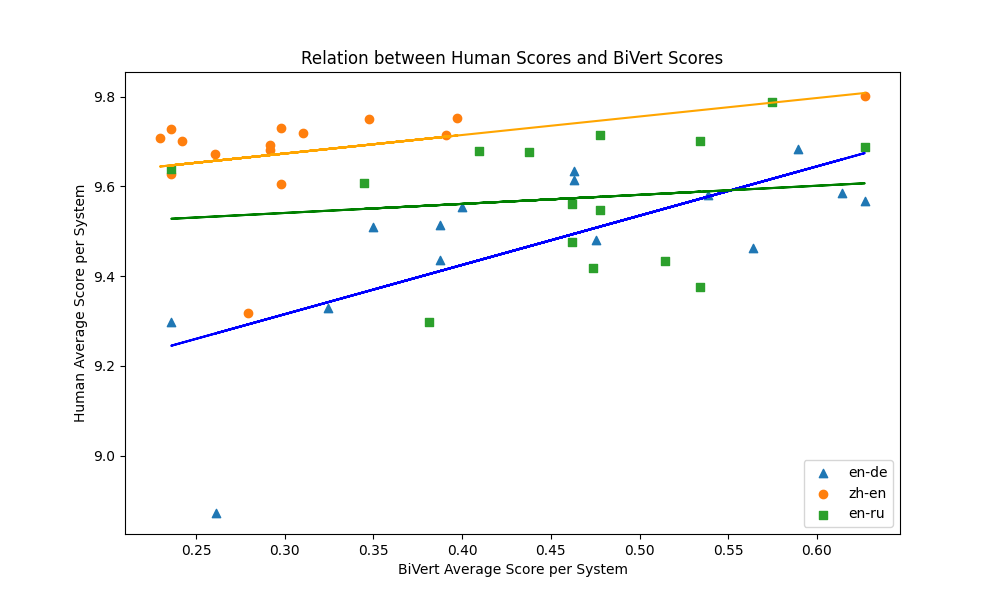} 
\caption{A comparison of average human scores and average \bivert{} scores for each language pair on all translation systems.}
\label{fig:correlation}
\end{center}
\end{figure*}

\section{Experiments}
In this section we describe the experiments conducted for finding the optimal \bivert{} configurations. For each language we learn the optimized values for \bivert{} features as resulted in \autoref{tbl:importances}.
For our self-supervised method, we start by applying a machine translation system on the source sentences to generate the back-translation.
We use a state-of-the-art translation model, MarianNMT~\cite{junczys-dowmunt-etal-2018-marian}, for this task.
This model is based on the Marian open-source tool for training and serving neural machine translation.
It was trained on multiple sources from parallel data collected at OPUS~\cite{TiedemannOPUS2004}. The model used the SentencePiece Tokenizer, an unsupervised text tokenizer, along with pre-trained embeddings from Word2Vec vectors~\cite{kudo-richardson-2018-sentencepiece}. Next, we make sure to apply a pre-processing language-specified routine on all data for optimal results.
For Chinese, we keep only Chinese characters in the text.
For English, we lowercase the sentence and expand contractions, for example \textit{don't} $\rightarrow$ \textit{do not}.
After cleaning the text we combine embeddings using the pairwise token technique.
The next step is aligning the words between the source sentence $s$ and its back-translated counterpart $s'$. 
For this process we calculate the score between each word pair $(w_1, w_2)$ as
$similarity = cos\_sim(w_1, w_2)$, using their embedding representations summed before. Since the algorithm searches for the minimum total cost we update each value to be $similarity = 1-similarity$.
For each aligned pair, we define the match relation and sum its value according to the category cost definition.
Specifically for the \textit{Sense} relation we apply Lemmatization\footnote{Simplemma: a simple multilingual lemmatizer for Python at https://github.com/adbar/simplemma} (currently only in English) prior looking up the words on BabelNet for accurate results. 
We restricted the sense connection edges to hypernym type only, and limited the graph depth to seven levels.
We operate on the most recent version 5.2  of BabelNet as our multilingual encyclopedia resource for extracting words' senses.
Finally, we learn the most optimal feature values to aggregate the summed up costs for each relation category, according to the original human reference scores in training data per sentence.
We learn the feature values by training a Gradient Boosting Regression model for each language pair.
Our training datasets are WMT Metrics Task MQM 2021 datasets in the following language pairs: English \textrightarrow{} German, English \textrightarrow{} Russian, and Chinese \textrightarrow{} English.
After fine-tuning our final model, we test our new evaluation method on the WMT Metrics Task MQM 2022 datasets for the same languages.
We compare our results to other evaluation techniques by calculating Pearson's correlation coefficient on the averaged human scores and averaged \bivert{} scores by translation system detailed in \autoref{app:eval}.

\subsection{Training}
We trained our model using Gradient Boosting Regression~\cite{friedman2002stochastic}, with different hyperparameters for each language pair. For both English \textrightarrow{} German and English \textrightarrow{} Russian we set the learning rate to 0.1; For English\textrightarrow{}German we used 100 estimators and max depth 6; For English\textrightarrow{}Russian we used 550 estimators and max depth 7.
Both data set labels, the human scores per sentence, are normalized for optimal training. Specifically in Russian training data, we normalized negative human scores to zero, as explained in~\cite{fonseca-etal-2019-findings} section 2.2.
For Chinese\textrightarrow{}English, we use 1000 estimators, max depth of 6, and set the learning rate to 0.05.
The English stopwords list is provided by NLTK,\footnote{Natural Language Toolkit \url{https://www.nltk.org/index.html}} and the Chinese stopwords list is from the Stopwords-iso library.\footnote{A collection of stopwords for multiple languages. \url{https://github.com/stopwords-iso/stopwords-iso}}
\autoref{tbl:sentences} displays the number of sentences used for training and predicting.

\begin{table}[!ht]
\centering
\begin{tabular}{lcc}
    \toprule 
     & \textbf{Train} & \textbf{Predict}\\
    \midrule
    \multirow{1}{*}{English-German} & 19,501 & 19,725 \\
    \multirow{1}{*}{English-Russian} & 12,000 & 19,725 \\
    \multirow{1}{*}{Chinese-English} & 16,124 & 28,124 \\
    \bottomrule
\end{tabular}
\caption{\label{tbl:sentences} Number of sentences used for training and prediction for each language pair. Prediction is for the whole WMT Metrics Dataset 2022 provided.}
\end{table}

\subsection{Results}
We evaluate how \bivert{}'s quality judgments fare in comparison to human scores on the full WMT Metrics Task 2022 Dataset.
The feature importance scores by language pair are presented in \autoref{tbl:importances} by language pair.
We evaluated our results by calculating Pearson's correlation between source-language \bivert{} average scores per system and the human gold-standard aligned scores.
We notice that for Chinese the Inflection and Derivation weights are zero, as these processes do not occur at the word level in Chinese.
We see that the Sense category is identified with the highest importance value in all language pairs. Thus indicating the success of BabelNet's sense network in assisting with the evaluation quality of the direct translation.
In \autoref{tbl:corr}, we compare our method scores with the correlation scores of other methods mirrored from the WMT22 Metrics Task findings.
Morever, \autoref{fig:correlation} represents a graphical display of the correlations calculated for \bivert{} in \autoref{tbl:corr}.
\bivert{} achieves the highest score for the English--German language pair among reference-less methods, as well as higher than BERTScore's.
For English--Russian \bivert{} achieves a middle-ranked score, and in Chinese--English the rank is lower.
This is perhaps due to the existing categories in use. It's possible that revising these categories to align better with the unique linguistic properties of the Chinese language could improve the results. Furthermore, the quality and coverage of BabelNet data for Russian and Chinese might play a significant role in the challenges we're observing.

\section{Conclusion and Future Work}
In this paper we present \bivert{}, a new multilingual reference-less method for evaluating machine translation.
This technique introduces an aspect of evaluation using graph senses extracted from semantic graphs, offering an untapped use case for these resources that is simple to implement and has immediate potential to achieve high results compared with human evaluation.
Its reference-free application mode allows high-quality evaluation of translation without need for parallel corpora, which can greatly lower the barrier for development of MT systems for low-resource languages and language pairs.

In the future, we aim to assess \bivert{}'s potential to be implemented in other generative NLP tasks. An additional avenue involves the potenital role switch between the evaluated system and the state-of-the-art system, where the evaluated system would back-translate the target sentence, thereby enchancing the consistensy of evaluations accross different systems. 
Moreover, we plan to expand our language categories to cover linguistically diverse languages, and also expand our graph knowledge of senses using resources other than BabelNet, such as Wikionary.
Finally, our word alignment algorithm does not currently deal with phrases or idioms, a fascinating avenue for future development.

\section*{Acknowledgments}

We thank the reviewers for their valuable comments.
This research was supported by grant no. 2022215 from the United States---Israel Binational Science Foundation
(BSF), Jerusalem, Israel.

\section*{Bibliographical References}\label{sec:reference}

\bibliographystyle{lrec-coling2024-natbib}
\bibliography{LREC/anthology, LREC/languageresource, LREC/lrec-coling2024-example, LREC/wsdmt}

\newpage

\appendix

\section{Individual evaluation of translation systems}
\label{app:eval}

We present the full evaluation scores for the systems in \autoref{tab:all-scores}.

\begin{table*}[!ht]
\centering
\begin{tabular}{llrr}
    \toprule
     & \textbf{System} & \textbf{Human } & \textbf{\bivert{}} \\
    \midrule
    \multirow{15}{*}{English \textrightarrow{} German} & bleu\textunderscore bestmbr & 9.615 & 0.614 \\
    & bleurt\textunderscore bestmbr & 9.555 & 0.609 \\
    & comet\textunderscore bestmbr & 9.567 & 0.627 \\
    & JDExploreAcademy & 9.581 & 0.620 \\
    & Lan-Bridge & 9.435 & 0.608 \\
    & M2M100\textunderscore1.2B-B4 & 8.872 & 0.598 \\
    & Online-A & 9.514 & 0.608 \\
    & Online-B & 9.585 & 0.626 \\
    & Online-G & 9.510 & 0.605 \\
    & Online-W & 9.684 & 0.624 \\
    & Online-Y & 9.480 & 0.615 \\
    & OpenNMT & 9.329 & 0.603 \\
    & PROMT & 9.297 & 0.596 \\
    & QUARTZ\textunderscore TuneReranking & 9.462 & 0.622 \\
    & refB & 9.634 & 0.614 \\
    \midrule
    \multirow{15}{*}{English \textrightarrow{} Russian} & bleu\textunderscore bestmbr & 9.715 & 0.296\\
    & comet\textunderscore bestmbr & 9.677 & 0.286 \\
    & eTranslation & 9.417 & 0.295 \\
    & HuaweiTSC & 9.476 & 0.292 \\
    & JDExploreAcademy & 9.679 & 0.279 \\
    & Lan-Bridge & 9.639 & 0.236 \\
    & M2M100\textunderscore 1.2B-B4 & 9.298 & 0.272 \\
    & Online-A & 9.561 & 0.292 \\
    & Online-B & 9.701 & 0.310 \\
    & Online-G & 9.687 & 0.333 \\
    & Online-W & 9.789 & 0.320 \\
    & Online-Y & 9.608 & 0.263 \\
    & PROMT & 9.548 & 0.296 \\
    & QUARTZ\textunderscore TuneReranking & 9.375 & 0.310 \\
    & SRPOL & 9.434 & 0.305 \\
    \midrule
    \multirow{15}{*}{Chinese \textrightarrow{} English} & AISP-SJTU & 9.682 & 0.443 \\
    & bleu\textunderscore bestmbr & 9.701 & 0.435 \\
    & bleurt\textunderscore bestmbr & 9.749 & 0.452 \\
    & comet\textunderscore bestmbr & 9.714 & 0.459 \\
    & HuaweiTSC & 9.692 & 0.443 \\
    & JDExploreAcademy & 9.718 & 0.446 \\
    & Lan-Bridge & 9.753 & 0.460 \\
    & LanguageX & 9.727 & 0.434 \\
    & M2M100\textunderscore 1.2B-B4 & 9.318 & 0.441 \\
    & Online-A & 9.627 & 0.434 \\
    & Online-B & 9.729 & 0.444 \\
    & Online-G & 9.707 & 0.433 \\
    & Online-W & 9.605 & 0.444 \\
    & Online-Y & 9.672 & 0.438 \\
    & refB & 9.801 & 0.497 \\
    \bottomrule
\end{tabular}
\caption{Individual system scores from WMT on human evaluation and through \bivert{}.}
\label{tab:all-scores}
\end{table*}

\end{document}